\documentclass[letterpaper, 10 pt, conference]{ieeeconf}  % Comment this line out if you need a4paper

\IEEEoverridecommandlockouts                              % This command is only needed if
\let\labelindent\relax

\usepackage{listings}
\usepackage{amsthm}

\usepackage{enumitem}
\usepackage{amsmath}
\usepackage{amssymb}
\usepackage{breqn}
\usepackage{mathtools,algorithm}
\usepackage[noend]{algpseudocode}
\usepackage{textcomp}
\usepackage{gensymb}
\usepackage{graphicx}
\usepackage{color}
\usepackage{varwidth}
\usepackage{microtype}
\usepackage{acronym}
\usepackage{multirow}
\usepackage{cite}
\usepackage{xcolor}
\usepackage[hidelinks]{hyperref}

\usepackage[noend]{algpseudocode}
\usepackage[symbol]{footmisc}
\usepackage[mathscr]{euscript}
\title{\LARGE \bf
Towards Generalized Robot Assembly through \\Compliance-Enabled Contact Formations
}

\author
{Andrew S. Morgan, Quentin Bateux, Mei Hao, and Aaron M. Dollar
\thanks{This work was supported by the United States National Science Foundation under Grants IIS-1752134, IIS-1900681, \& NRI-1734190.}
\thanks{Authors are with the Department of Mechanical Engineering and Materials Science, Yale University, USA. (email: {\tt\footnotesize \{andrew.morgan, quentin.bateux, mei.hao, aaron.dollar\}@yale.edu}).} 
}

\begin{document}

\maketitle
\thispagestyle{empty}
\pagestyle{empty}

%%%%%%%%%%%%%%%%%%%%%%%%%%%%%%%%%%%%%%%%%%%%%%%%%%%%%%%%%%%%%%%%%%%%%%%%%%%%%%%%
\begin{abstract}

Contact can be conceptualized as a set of constraints imposed on two bodies that are interacting with one another in some way. The nature of a contact, whether a point, line, or surface, dictates how these bodies are able to move with respect to one another given a force, and a set of contacts can provide either partial or full constraint on a body's motion. Decades of work have explored how to explicitly estimate the location of a contact and its dynamics, e.g., frictional properties, but investigated methods have been computationally expensive and there often exists significant uncertainty in the final calculation. This has affected further advancements in contact-rich tasks that are seemingly simple to humans, such as generalized peg-in-hole insertions. In this work, instead of explicitly estimating the individual contact dynamics between an object and its hole, we approach this problem by investigating compliance-enabled contact formations. More formally, contact formations are defined according to the constraints imposed on an object's available degrees-of-freedom. Rather than estimating individual contact positions, we abstract out this calculation to an implicit representation, allowing the robot to either acquire, maintain, or release constraints on the object during the insertion process, by monitoring forces enacted on the end effector through time. Using a compliant robot, our method is desirable in that we are able to complete industry-relevant insertion tasks of tolerances $<$0.25mm without prior knowledge of the exact hole location or its orientation. We showcase our method on more generalized insertion tasks, such as commercially available non-cylindrical objects and open world plug tasks.

\end{abstract}

%%%%%%%%%%%%%%%%%%%%%%%%%%%%%%%%%%%%%%%%%%%%%%%%%%%%%%%%%%%%%%%%%%%%%%%%%%%%%%%%
\section{Introduction}

%%Robot manipulation and assembly tasks, generally

% Unstructured worlds are full of uncertainty. 
% The physical world is made up of matter and between all matter there exists some sort of interaction. For robots, this interaction typically exists in the form of physical contact between bodies. 
Robots interact with the world through physical contact. Contact has been studied in the literature for decades and can be conceptualized in a number of ways, but in its most fundamental sense, contacts are interaction-based constraints imposed on the relative motion between objects. As humans, we often rely on contact to change the state of our environment, whether it be removing keys from a bag or cleaning a dirty dish. While seemingly simple, contact is quite complex in that it requires accurate parameter estimation in order to sufficiently predict its current state, i.e., static, stick-slip, or slip conditions. Previous works have tried to estimate various properties of contact during manipulation, e.g., the location of the contact \cite{ bimbo2022force}, frictional and curvature properties \cite{TrinkleContact,Montana1992}, slip conditions \cite{brock1988enhancing}, modes \cite{morgan2021TRO}, etc., but have witnessed many bottlenecks. 

\begin{figure}[thpb]
      \centering
      \includegraphics[width=0.4\textwidth]{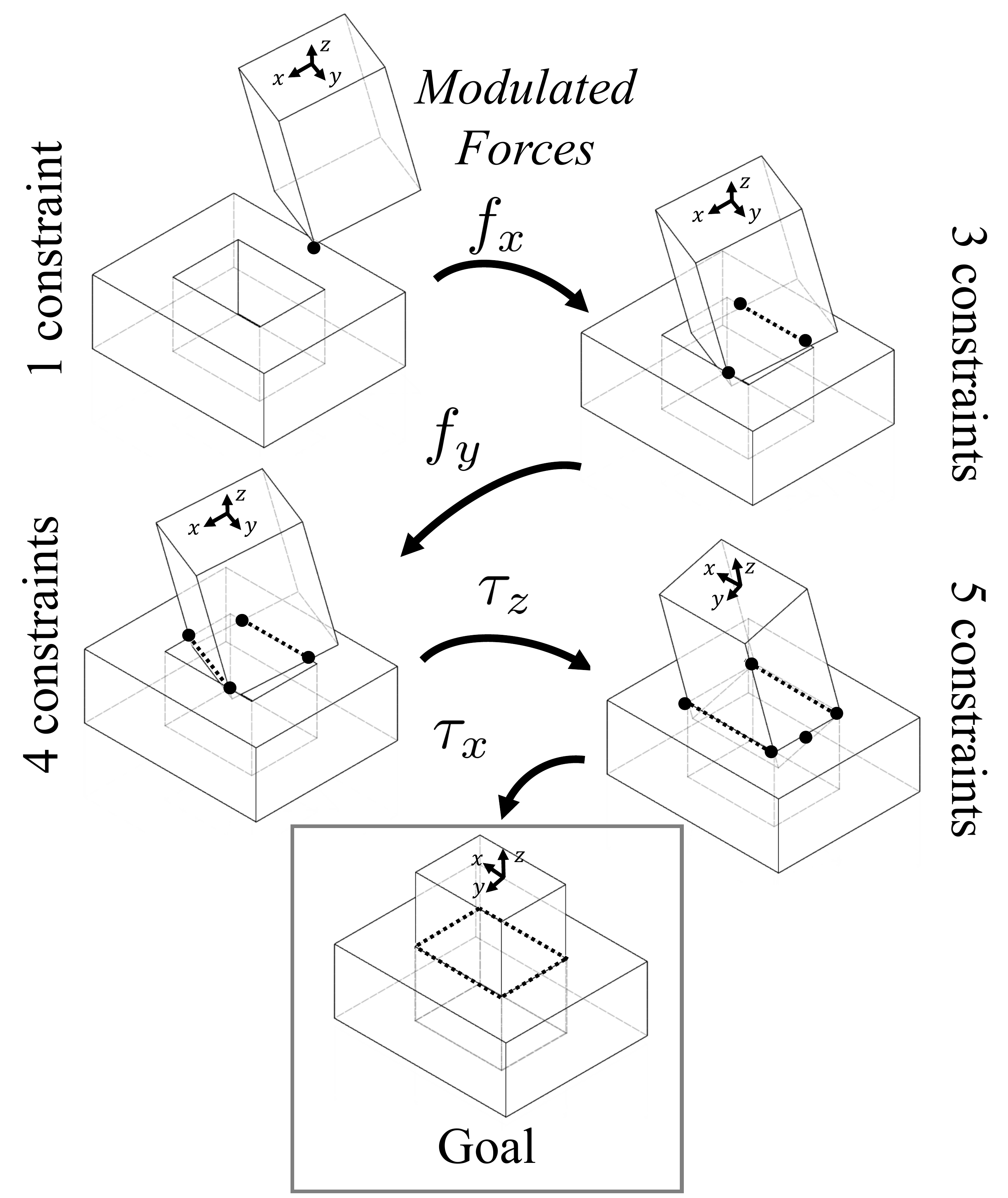}
      \vspace{-0.1cm}
      \caption{Object insertion can be conceptualized as the continual addition and modulation through time of an object's constrained degrees-of-freedom. By continually modulating forces once constraints are detected, tight tolerance insertion can be achieved without \textit{a priori} knowledge of the object geometry or exact hole pose for convex objects.}
      \label{Splash}
      \vspace{-0.65cm}
  \end{figure}

%Notably, the robotics community has been continually studying physical interactions as the next generation of robots must be able to easily operate in unstructured and contact-rich environments alongside humans. 

Fundamentally, contacts define constraints and constraints define the available degrees-of-freedom (DOF) of an object. In free space, an object is able to move in 6-DOF, but a non-cylindrical peg inserted into a slot is only able to translate in 1-DOF. When there exists some degree of uncertainty in the estimation of contact, the calculated constraints can be non-representative of the system's true state. For instance, let's assume there exists little uncertainty in the estimation of point contact locations. Here, we can claim that the restricted motion of an object with two point contacts may be similar or even identical in nature to the constraints imposed on an object with 100 edge point contacts on or near a single axis. In a practical sense, this distinction is insignificant in the free DOFs of the object, and merely adds needless computation. Now, let's be more practical and assume there does exist some degree of uncertainty in our estimation of contact locations, even in a single point out of the 100. This calculation may estimate additional constraints placed on the object which are not physically valid. Thus, instead of explicitly defining the state of every contact between a robot and its environment, which is quite computationally expensive and is gravely subject to uncertainty, we in this work abstract out individual contact properties and are conversely interested in contact formations, or more generally, the constraints placed on an object's motion when in contact with its environment.

\textit{Contact formations} can be conceptualized as a grouping of contact sets that constrain an object in similar ways \cite{skubic1996identifying}. For instance, various contact scenarios could constrain either some translational DOFs, e.g., sphere in tube, or all translational DOFs, e.g. sphere in case. A set of all contact groupings that constrain an object in a similar way would be considered in a contact formation. Traditionally, acquiring and maintaining contact formations on a real robot platform was difficult due to an inability to practically modulate forces between two bodies over time. Within the past decade, a great deal of work has addressed the concept of dealing with uncertainty in robot mechanics, specifically, the ability to adapt to external forces as to continually maintain contact with an object. Compliant manipulation frameworks, either in the form of active \cite{Park2017} or passive architectures \cite{polverini2016, choi2016vision}, enable this ability -- allowing a system to kinematically adapt to uncertain environments and maintain desired forces. This introduces the idea of \textit{compliance-enabled contact formations}, which we leverage as to maintain a desired type of contact during the task. 

Combining these ideas, this work investigates the applicability of controlling tight tolerance insertion tasks via the chaining of contact formations. Specifically, we leverage a control approach that modulates forces in controlled directions, and by continually adding forces along specific axes until contact, i.e., constraints, we transition between different contact formations until the task is successfully completed (Fig. \ref{Splash}). This approach utilizes external contact as a means of constraining the object's potential motion \cite{Chavan-Dafle-2014-7860}, which in turn limits uncertainty of the object's current configuration. While utilizing a compliant robot, this system aids in both acquiring and maintaining a desired contact configuration. Moreover, we show how in-hole jamming is limited due to this compliance, functioning similarly in concept to a remote center of compliance (RCC) device \cite{xu1990robot,ciblak1996remote}. Retrofitted with a 6-axis force/torque sensor at the end effector and an in-hand camera, we monitor the state of the object and servo to its desired goal forces and axial alignment.
% Once all constraints are added accordingly, the task is completed. 
We can thus summarize our contributions:
\begin{enumerate}
  \item Our algorithm is an object-agnostic procedure that does not require \textit{a priori} knowledge of the object geometry, exact hole position, or exact hole orientation. To our knowledge, this is one of few works that can insert an object when axial rotation is largely unknown ($>40\degree$). 
  \item We do not require complex analytical or computationally expensive learned models of contact dynamics; our method relies solely on the idea of validating constraints via axial force application through time.
  \item This method is conceptually straightforward, as it solely relies on following a trajectory in Contact Formation-space, and can thus be controlled physically in real-time.
  \item We justify the true utility of using compliant systems for tight tolerance, contact-rich tasks by quantifying reconfigurability. Specifically, we quantify slip conditions for our underactuated hand, and show how the hand reorganizes its contacts when excessive forces are applied, which helps eliminate in-hole jamming. 
\end{enumerate}
% The continuation of this paper is organized as follows: Sec. II introduces ... 

\section{Related Work}
Generalized robot assembly has been investigated for decades and the development of a practical, holistic solution has faced many challenges \cite{kemp2007}. Elucidating this grand challenge can be found in previous works with: standard cylinders \cite{Park2017,tang2016,tang2016learning}, multiple-peg objects \cite{xu2019, fei2003}, soft or compliant cylinders \cite{zhang2019}, industrial inserts \cite{schoettler2020}, and standard open world objects \cite{polverini2016,levine2015learning,choi2016vision,chang2018robotic,she2020cable}. In an attempt to extend previous work, we outline a method that underscores the idea of \textit{generalization}, in that our method provides a practical solution to insertion when there exists positional and orientational uncertainty of the hole pose, with minimal knowledge of the object geometry. 

\textbf{Peg-in-hole Assembly:} Approaches to solving generic assembly tasks can be divided into two different categories: analytical approaches and learning-based approaches. In the former, approaches utilize contact models to reason about their state conditions \cite{fei2003, zhang2019}, controlling the manipulator based on force/torque sensor readings \cite{tang2016}. Model-based approaches using vision and rigid systems have been difficult to implement, as uncertainty in the manipulator's pose can create dangerously large forces \cite{xu2019compare}. Compliance has become a key component to overcoming this issue,
% for modeling, sensing, and control, 
both in software-based \cite{Park2017} and hardware-based \cite{polverini2016, choi2016vision} architectures. 
% Though a majority of solutions 
Many works have disregarded the use of a dexterous robot hand, directly attaching the object to the manipulator. Two previous works utilized a dexterous hand \cite{vanwyk2018,Morgan2021RSS}, where both found it advantageously extended the robot's workspace. In both cases, \textit{a priori} knowledge of the hole pose was available. 

Learning-based solutions help deal with robot sensor and model uncertainty, and typically require physical environment exploration or some form of human-in-the-loop demonstrations \cite{tang2016learning}. Reinforcement learning and self-supervised learning \cite{lee2019making, zakka2020form2fit} have been popular approaches, allowing the robot to interact with its environment and sufficiently explore desired regions \cite{schoettler2020,levine2015learning,Beltran_Hernandez_2020}. Notably, learning manipulation policies is both, time consuming and data intensive, which increases the chance of damaging the robot. 

While attempting to maintain a model-free nature but also limiting the need for expensive exploration, we are interested in forming a solution that can achieve reliable insertions of tight tolerance by solely using force data and vision feedback, and without \textit{a priori} knowledge of the hole pose.

\textbf{Contact Formations: }
\textit{Contact formations} ``provide a qualitative description of how 2 or more objects make contact with one another (e.g., vertex to surface, edge to edge)" \cite{skubic1996identifying}. This formulation is advantageous, in that it implicitly defines the contacts and thus constraints imposed on an object's motion. Contact formations have been computed in different ways: from using CAD models \cite{Thomas2011} to probabilistic frameworks based on interaction \cite{farahat1995}. From knowing these contact formations, other works have utilized them in planning \cite{cheng2021contact} and control frameworks \cite{cabras2010contact}. Fundamentally, the ability to group together different combinations of contacts that define the same or mere similar constraints, can be a powerful tool for defining more capable robot manipulation capabilities.   

\begin{figure*}[ht]
\centering
\includegraphics[width = 0.78\textwidth]{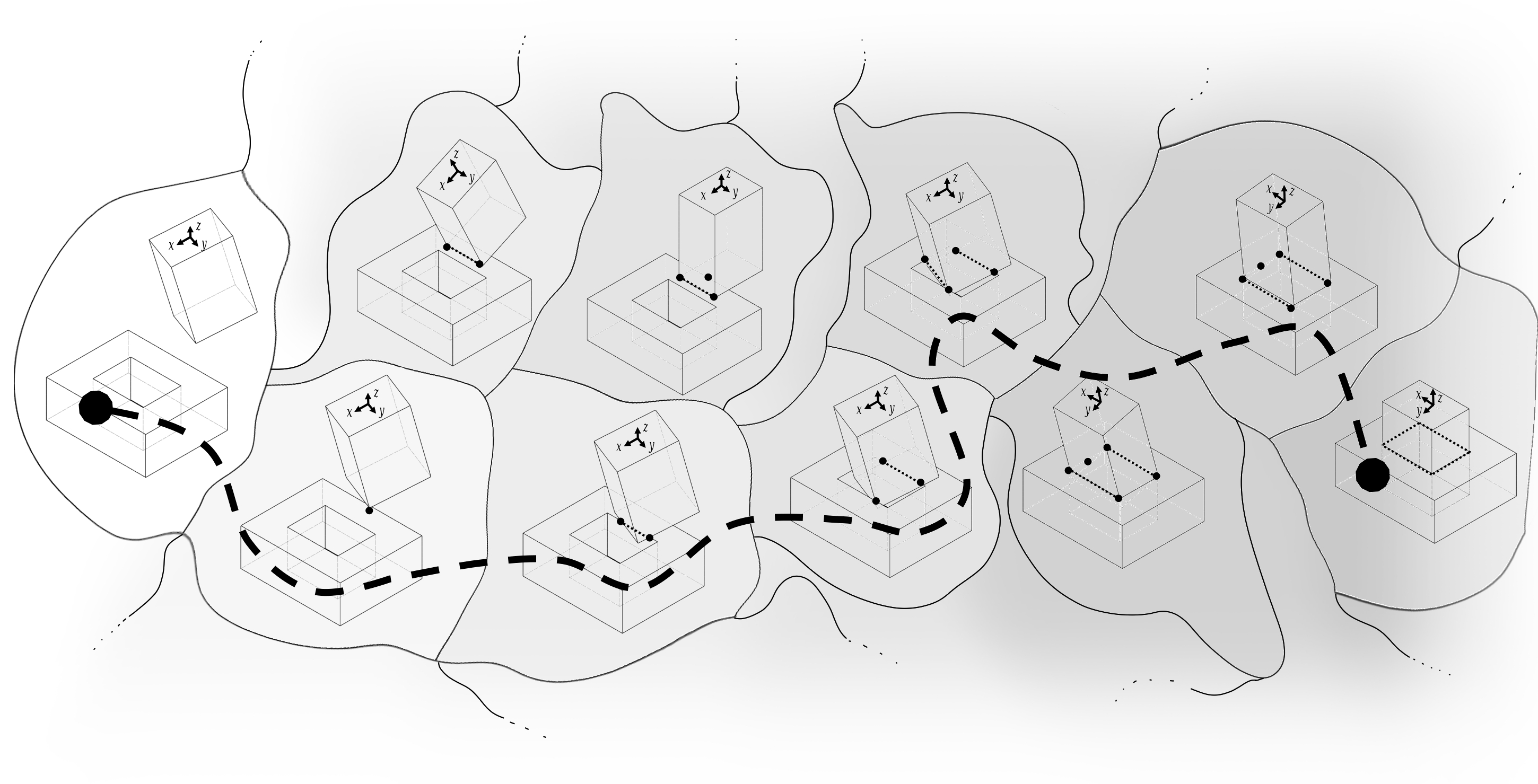}
\vspace{-0.25in}
\caption{The Contact Formation-space (CF-space) of an insertion task can be conceptualized as an explicit organization of different contact formations that share borders according to constraint similarity and possible transitions. Transitions between contact formations are represented when constraints are added or removed to the state of an object. By controlling paths through a CF-space from light (few constraints) to dark (more constraints), the object follows a progression towards insertion. \textbf{Note: }This depiction of contact types is not exhaustive and other intermediate formations may be possible. }
\label{fig:Concept}
\vspace{-0.2in}
\end{figure*}

\section{Methodology} \label{sec_methods}

%Thank you Quentin for writing some of this. Thoughts in general:
%- We could follow something similar to this https://arxiv.org/pdf/2110.03555.pdf

%-There will have to be some definitions in there, of what we mean by a contact formation, etc. I have them somewhat defined in the above sections, but we may have to write it once again. 

%-I saw in that old paper that they call our Fig. 2 a "Contact Formation Space" or CF-Space. I dont mind rolling with that sort of terminology. The hard part will be defining the dimensionality/shape of this space. We can either look to them for inspiration or so something ourselves. https://link.springer.com/content/pdf/10.1007/978-3-642-83625-1.pdf on page 367

%-I am not sure if we will need another figure or not, but if we do, please feel free to make one and insert where necessary. We will trim everything at the end. 

%Follow something similar to this
%https://arxiv.org/pdf/2110.03555.pdf

% Say we want to use contact formations, but do it because we have a compliant system. So we do "compliant contact formation"

We are interested in solving a generalized peg-in-hole problem by leveraging \textit{compliance-enabled contact formations} -- the concept that object constraints can be more easily acquired, broken, and remade when operating within a compliant robot's ``reconfiguration range" \cite{hang2019pregrasp, morgan2022}. Conceptually, compliance allows the robot to convert a traditionally difficult force control problem into a velocity control problem, as there now exists an ability to ``take up the slack" in control uncertainty. A F/T sensor is leveraged to modulate forces closed-loop. This is particularly advantageous for maintaining a contact formation, as now we can more easily ensure the system maintains constraints while operating along a desired path. 

Fundamentally, contact formations (CF) represent groupings of discrete positional and physical relationships. Solving the insertion problem can thus be framed as a traversal from a starting, $CF_0$, toward a final, $CF_n$, transitioning through intermediate CFs, i.e., $CF_{path}=\{CF_0, CF_1,\dots,CF_{n}\}$ (Fig. \ref{fig:Concept}). In this work, we demonstrate the efficacy of this approach by following a fixed path in CF-space.% that allows us to insert a variety of objects.

Formally, transitioning from one CF, $CF_n$, to a new CF, $CF_{n+1}$, along a trajectory in CF-space is comparable to adding or removing constraints on the object's available DOF. 
% Notably, operating within a single CF does not mean that the relative pose between the object and the hole is the same, but actually that the constraints imposed on the object's DOF are.
Thus, the role of compliance serves to ensure that a single desired CF is maintained while manipulation occurs via force modulation, and transitions are detected when the \textit{motion of the object is impeded by the environment}. Once a variational change in F/T signals is felt by the robot, a contact formation change is detected, and the system has transitioned in CF-space. Conceptually, \textit{move along a single axis until no longer possible, and while maintaining that force, move orthogonally until another constraint is added.}

The theory of controlling compliance-enabled contact formations is that constraints can be continually added until the task of insertion is complete. Notably, any DOF that is not currently constrained can be controlled to perform other tasks without modifying the current CF of the system. For example, an edge contact between an object and the hole's surface allows the robot to explore via sliding until a lateral force is detected and thus a new CF is achieved. It is important to note that under this additive force contact process, explicitly estimating the current CF is possible if the starting CF is known and the transitions can be detected via F/T signals.

\subsection{Assumptions and Prior Knowledge}
By relying on F/T sensing for environment perception, our algorithm leverages the following priors:
\begin{itemize}
    \item The object starts in a stable, centered, and upright grasp. This grasp is expected to be maintained.
    \item $CF_0$ has 0 constraints, i.e. 6 free DOFs starting in space.
    \item The hole is somewhere within a known 2D workspace boundary in the XY-plane (e.g., 10cm exploration size). 
    \item Minimal knowledge of the object's face type is known, but assumed to be convex or near-convex.
    \item There exists a hole with low contact friction and positive tolerance matching the peg within the workspace.
    \item Within the workspace, no objects other than the hole, plane, and manipulated object are present. This ensures the CF transitions match the preset CF-space path.
\end{itemize}

\subsection{CF-space Insertion Algorithm}

Our approach follows a desired path in CF-space, as defined through a total of 7 steps. For ease of notation, we set the object frame as an orthogonal frame with the $x-axis$ pointing toward the direction of the object motion on the hole surface plane, and the $z-axis$ pointing downward. For each step, we provide a corresponding implicit CF control target. Although we define a set of values for force modulation, this is for clarity and is in practice robot-specific (see Sec. \ref{sec:forcePlateaus}). 

% is for additional clarity and corresponds to an idealized rectangular-faced object, as illustrated in Fig.~\ref{Splash}.

Let's assume we can detect the object forces, $F=\{f_x, f_y, f_z\}$ and torques, $T = \{\tau_x, \tau_y, \tau_z\}$, during insertion. Our goal is for the robot to traverse through a desired trajectory of CFs $\{CF_0, \dots, CF_5\}$ (Fig.~\ref{Splash}). We outline the algorithm in pseudocode below, and provide a breakdown on a real robotic system in Sec. \ref{sec:tightTolerance}.
% Fig.~\ref{fig:forcePlot}\textbf{(a)(b)}.
\newpage
\vspace{-0.2in}
\hrule height 1pt depth 1pt width 0.48\textwidth \relax
\vspace{-0.05in}
\noindent \begin{center} \large{\textbf{Insertion Pseudocode}}\end{center}
\vspace{-0.02in}
\hrule height 1pt depth 1pt width 0.48\textwidth \relax
\vspace{-0.02in}
\begin{enumerate}[label=\textbf{\arabic*)}]
    \item 
    \textbf{Reaching the hole plane}\\
    \textit{Current constraints:} None\\
    \textit{Target constraint:} ($+f_z$, 1.5N)\\
    \textit{Implicit CF targets:} ($CF_1$) point or edge on face\\
    \textit{Additional motions: }
    \begin{itemize}
        \item Lateral exploration within the workspace area, randomly selecting x/y direction within workspace limits.
        \item In-hand manipulation to tilt the object toward the direction of the lateral motion, to ensure an edge/face contact, and avoid a face/face contact. 
        % (causes sliding over the hole under partial overlap)
    \end{itemize}
    
    \item
    \textbf{Searching for the hole}\\
    \textit{Current constraints:} ($+f_z$, 1.5N)\\
    \textit{Target constraint:} ($+f_x$, 0.7N)\\
    \textit{Implicit CF target:} ($CF_2$)  3-point contact with hole\\
    \textit{Additional motions:} Lateral exploration as previous step.
    
    \item
    \textbf{Wedging}\\
    \textit{Current constraints:} ($+f_z$, 1.5N), ($+f_x$, 0.7N)\\
    \textit{Target constraint:} ($+f_y$, 0.7N)\\
    \textit{Implicit CF target:} ($CF_3$) 4-point contact with hole
    
    \item
    \textbf{Rotational alignment of peg and hole}\\
    \textit{Current constraints:} ($+f_z$, 1.5N), ($+f_x$, 0.7N), ($+f_y$, 0.7N)\\
    \textit{Target constraint:} ($+\tau_z$, 0.1N/m)\\
    \textit{Implicit CF target:} ($CF_4$) hinge-type contact\\
    \textit{Note:} This step applies only to non-cylindrical objects. %objects with a non-circular face.
    
    \item
    \textbf{Correcting upward tilt}\\
    \textit{Current constraints:} ($+f_z$, 1.5N)\\
    \textit{Target constraint:} ($+f_x$, 0N), ($+f_y$, 0N)\\
    \textit{Implicit CF target:} ($CF_5$) Prismatic joint-type contact\\
    \textit{Additional motion:} Rotation around $x-$ and $y-axes$ to minimize the accumulated angle between the object and the hand due to fingertip slip during the previous steps. This rotation is performed both with in-hand manipulation and arm motions.\\
    \textit{Note:} The lateral forces are now minimized to avoid jamming the object. This also helps centering the object in the hole if the peg rotation is not perfectly centered on the object's center of mass. The angle information is provided by extracting the 3D pose of a marker placed on the surface of the object as seen from the palm camera.
    
    \item
    \textbf{Inserting peg}\\
    \textit{Current constraints:} ($+f_z$, 1.5N), ($+f_x$, 0N), ($+f_y$, 0N)\\
    \textit{Exit condition:} When the fingertips start touching the hole surface or the object hits the hole bottom, we switch to disengagement. This can be detected as a sharp increase in $f_z$.\\
    \textit{Note:} We have already reached the CF state that enables peg insertion, thus the system maintains it while the final free DOF, i.e., the $z-axis$ translation, is controlled to perform the final insertion motion.
    
    \item
    \textbf{Detaching hand grip and retracting arm}\\
    \textit{Action: }The hand opens and the arm returns to its origin position in an open loop motion.
\end{enumerate}
\vspace{-0.02in}
\hrule height 1pt depth 1pt width 0.48\textwidth \relax

\section{Experiments}

\begin{figure}
 \centering
 \includegraphics[width = 0.4\textwidth]{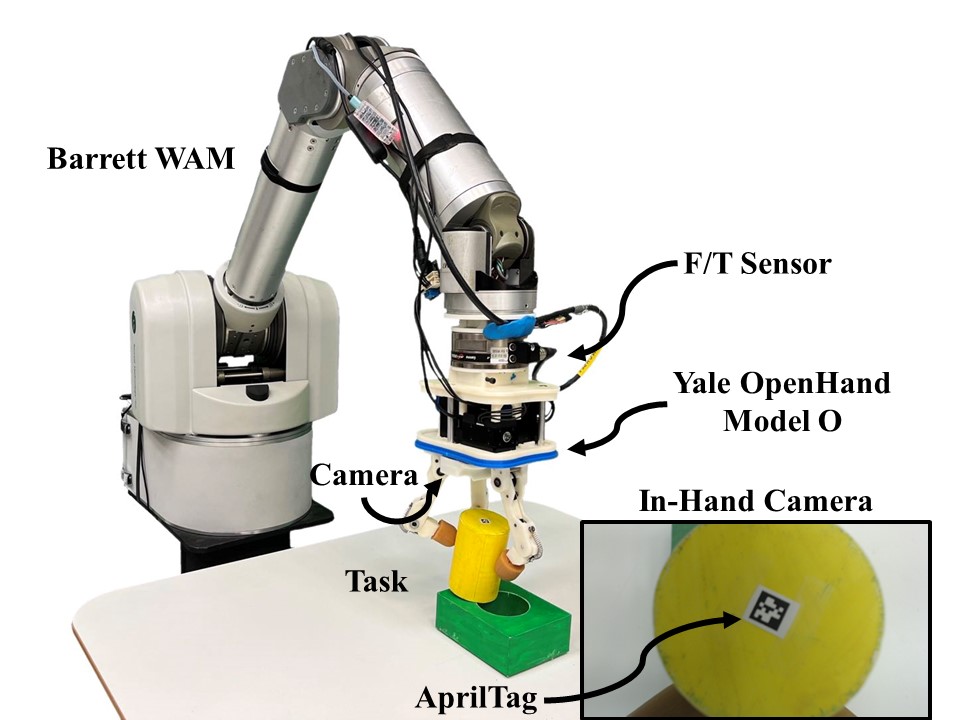}
 \caption{A Yale OpenHand Model O and a 6-axis force/torque sensor are affixed to the end of Barrett WAM manipulator. Inside of the palm of the hand, an in-hand camera setup is fabricated as to monitor the state of the object during manipulation via an AprilTag.  }
 \label{fig:System} 
 \vspace{-0.2in}
 \end{figure}

We evaluate our algorithm on a low-impedance manipulator and a compliant, underactuated hand (Fig. \ref{fig:System}). The manipulator, a 7-DOF Barrett WAM, utilizes the $RRTConnect$ algorithm via OMPL \cite{sucan2012open} for global planning and is controlled locally via a Jacobian-based velocity controller. The manipulator is imprecise due to an inaccurate internal model of its true system dynamics, which further challenges our algorithm's robustness. The end effector, an adapted Yale OpenHand Model O \cite{ma2017}, is a passively adaptive hand consisting of three actuators and six total joints. The hand is not equipped with tactile sensors or joint encoders, as to reduce weight and cost, making estimation of the true system state difficult during manipulation \cite{hang2020hand}. Modifications to the readily available open source design include rounded fingertips, as to facilitate in-hand manipulation, and bearings within the joints. An in-hand manipulation controller is devised, as in \cite{morgan2020}, and is utilized for fine motor control of object orientation up to $\pm20\degree$. Connecting the hand to the arm is a 6-axis ATI force/torque sensor sampled at 30Hz. Finally, a camera is fabricated into the palm of the hand as to enable the monitoring of object poses during manipulation via AprilTags.

\begin{figure}
 \centering
 \includegraphics[width = 0.48\textwidth]{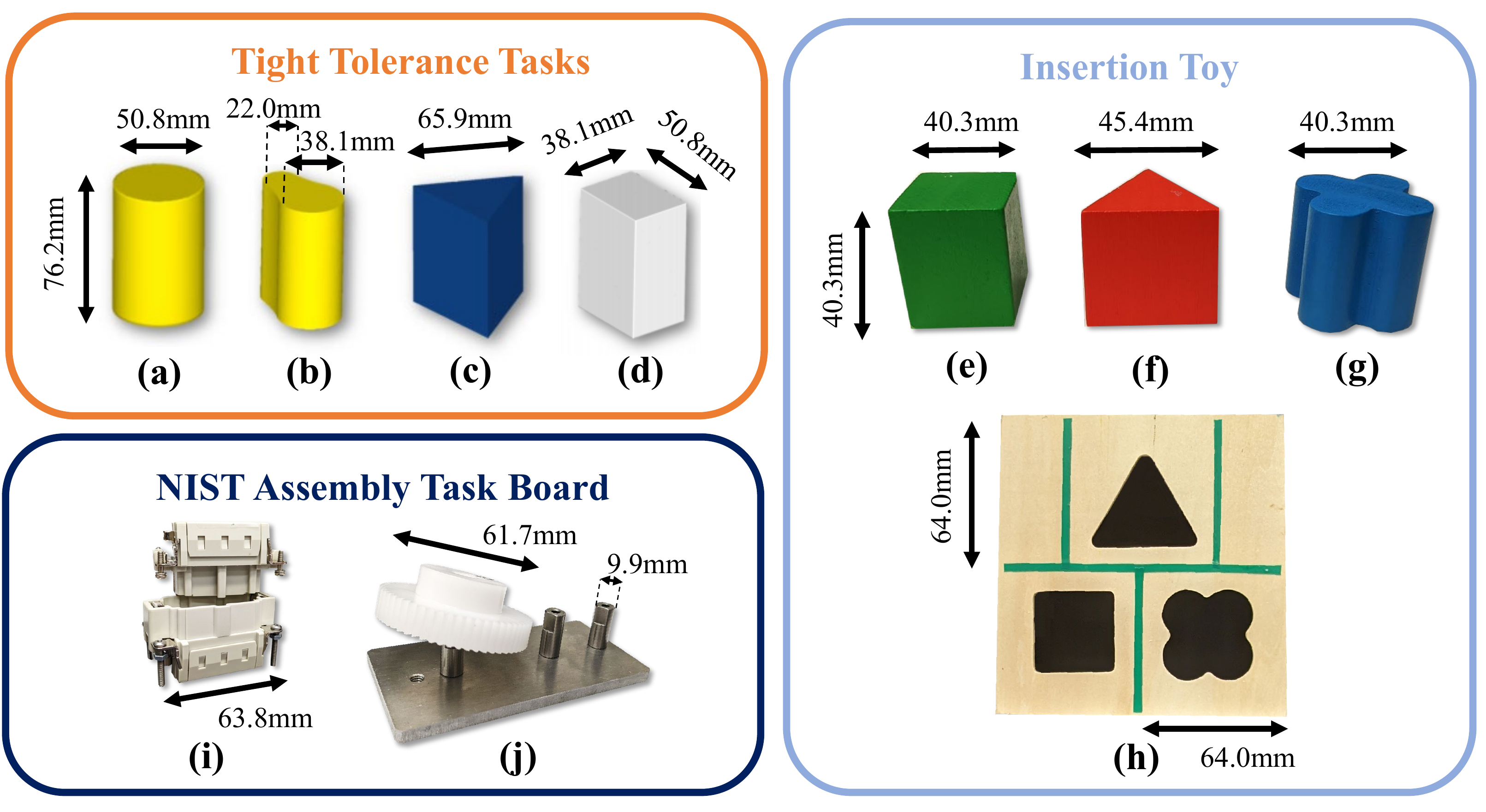}
 \caption{Tested objects can be classified into three categories: Tight Tolerance Tasks, Insertion Toy, and NIST Assembly Task Board. Objects are referenced according to their face geometries: \textbf{(a)} circle, \textbf{(b)} pear, \textbf{(c)} large triangle, \textbf{(d) }rectangle, \textbf{(e)} cube, \textbf{(f)} small triangle, and \textbf{(g)} clove. Subfigure \textbf{(h)} illustrates the insertion toy's hole layout with designated search spaces in green. NIST objects \textbf{(i)} and \textbf{(j)} are referred to as the plug and gear, respectively.}
 \label{fig:objs} 
\end{figure}

The algorithm is validated through a variety of experiments. In our first experiment, we develop a linear pusher, comprised of a leadscrew and a stepper motor, to displace objects along different axes of the gripper's workspace. During this test, we measure the forces exerted on the object by the pusher and note the amount of force the gripper can resist before fingertip slip occurs. Thereafter, we test our algorithm with tight tolerance tasks ($<$0.25mm), objects commercially available in a children's insertion toy, and two tasks within the NIST Assembly Task Board \#1 \cite{kimble2020benchmarking} (Fig. \ref{fig:objs}). 

%%%%%%%%%%%%%%%%%%%%%%%%
\subsection{Quantifying Axial Compliance} \label{sec:forcePlateaus}

\begin{figure}
 \centering
 \includegraphics[width = 0.40\textwidth]{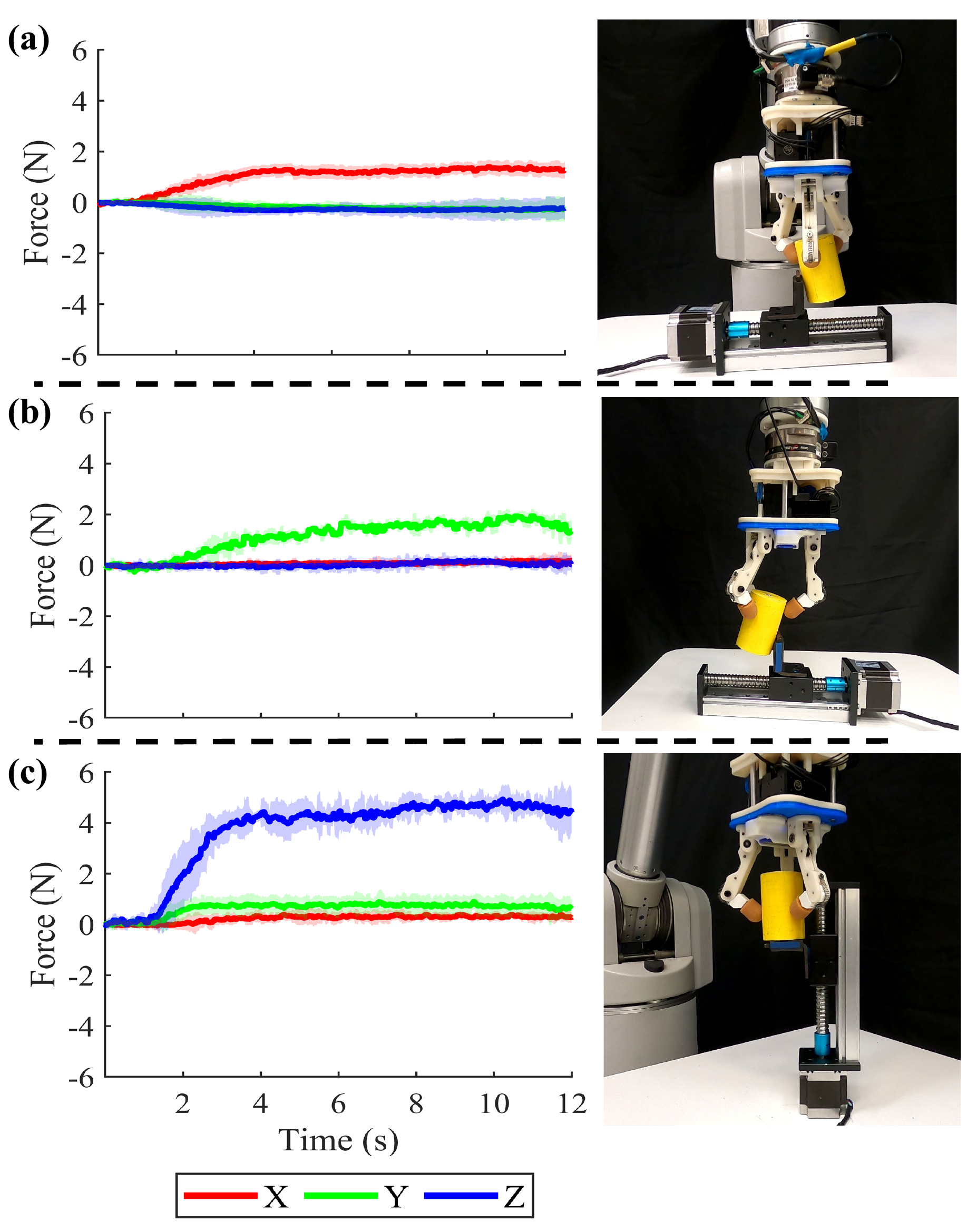}
 \caption{An object is pushed along the (a) $x-$, (b) $y-$, and (c) $z-axes$ with a linear pusher to evaluate the \textit{force plateau} along each dimension, or more specifically, the amount of force the hand can resist before slip occurs. }
 \label{fig:linearPush}
 \vspace{-0.2in}
\end{figure}

First, we are interested in quantifying the compliance of our system. Conceptually, a purely compliant system with fixed contact could comply indefinitely until a hard stop is reached, but practically for grippers, there exists a limit in which the forces applied to an object can be resisted by fingertip contacts. This information is valuable to quantify as it enables the system to predict when slippage may occur, which can in turn be used to inform contact formation switching. 

Given the linear pusher operating at a velocity of 3mm/s, we push the \textbf{(a)} circle grasped by the manipulator along the $x-$, $y-$, and $z-axes$ for a duration of 12 seconds and over six trials. After each trial, the system is systematically reset. We record the forces measured by the F/T sensor via the pusher during each trial, and plot the mean and standard deviation for each evaluation (Fig. \ref{fig:linearPush}). Note that for each linear push, a \textit{force plateau} is distinguishable, at $\sim$1.5N for the $x-$ and $y-axes$ (Fig. \ref{fig:linearPush}(a)(b)) and at $\sim$4N for the $z-axis$ (Fig. \ref{fig:linearPush}(c)). These plateaus correspond to the hand's ability to resist forces in corresponding directions, or more physically, the point at which static friction of the contact is overcome and a new hand-object configuration is realized. 
% As a contact slides, the hand passively reconfigures and a stable grasp is once again acquired thereafter. 

This data is valuable in that we are able to quantify the stable contact operating region given an external force. Once a force level is exceeded, sliding occurs and the hand-object state reconfigures until stability is once again realized. We can use these force plateaus for defining modulating forces for our algorithm in Sec. \ref{sec:tightTolerance}.    

\begin{table}[h]
    \caption{\textsc{\small{Metrics for Object Insertion Experiments}}}  
    \label{tab:main_insertion}
    \renewcommand{\arraystretch}{1.0}
    \centering
    \addtolength{\tabcolsep}{-3pt} 
     \begin{tabular}{c c c c c c}   
     &\textbf{Obj. }&\textbf{Tol. (mm)} &\textbf{Explore (s)}&\textbf{Insert (s)}&\textbf{Offset (\degree)}\\[0.4ex] 
     \hline\hline
     \textbf{(a)} &circle& 0.25  & 37.2 & 28.0 & $^*$n/a\\
    %  \hline
     \textbf{(b)} &pear& 0.25 & 31.7 & 94.7 & 41.1\\ 
    %  \hline
     \textbf{(c)} &l. triangle& 0.25  & 34.1 & 65.8 & 29.8 \\ 
    %  \hline
     \textbf{(d)}& rectangle& 0.25 & 44.2 & 48.1 & 27.1\\ 
     \hline
     \textbf{(e)}& cube& 3.0 & 27.3 & 30.8 & 29.2 \\ 
     \textbf{(f)}& s. triangle& 2.1 & 45.1 & 21.2  & 25.0 \\ 
     \textbf{(g)}& clove& 2.6 & 63.2 & 19.4  & 23.7\\ 
     \hline
     \textbf{(i)}& plug& 0.9 & 29.8  & 31.0  &  36.2 \\ 
     \textbf{(j)}& gear& 0.1 & 27.9  & 21.4 & $^*$n/a\\ 
     \hline
    \end{tabular}\\
    \begin{flushleft}\footnotesize{$^*$not applicable for pegs with circular faces}\end{flushleft}
    \vspace{-0.2in}
\end{table}

\subsection{Tight Tolerance Insertion} \label{sec:tightTolerance}
\begin{figure*}
 \centering
 \includegraphics[width =0.96\textwidth]{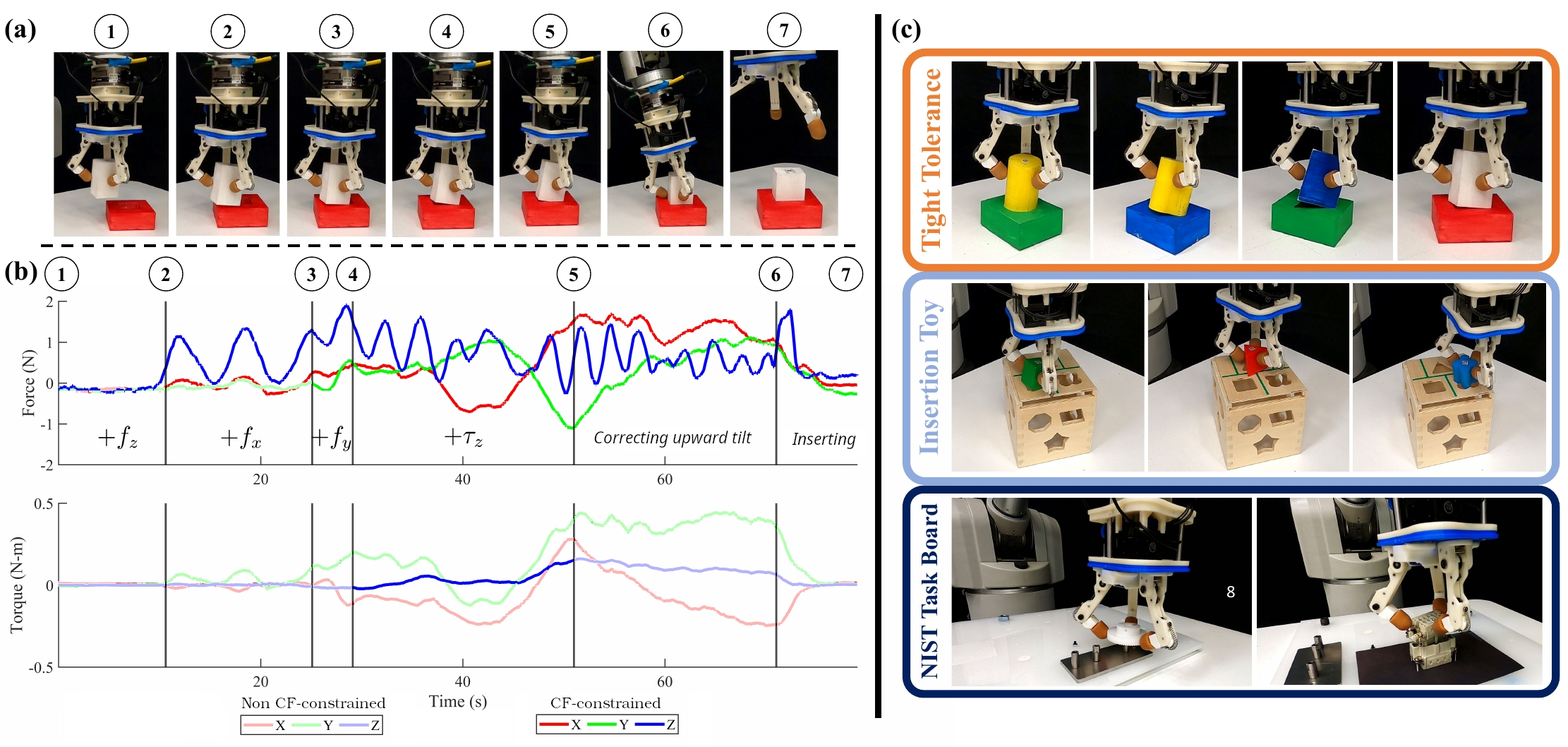}%{Figures/test.pdf}
 \caption{The progression of insertion is depicted in (a), where an object goes from free space (0 constraints) to inserted (5 constraints) into its goal configuration. (b) During this process, forces are modulated and added through different steps in the insertion task (forces are smoothed and placed in the world frame for clarity). (c) We evaluate this algorithm with tight tolerance tasks, child insertion toys, and objects from the NIST Task Board.  }
 \label{fig:forcePlot} 
 \vspace{-0.2in}
\end{figure*}

% \begin{figure}
%  \centering
%  \includegraphics[width = 0.40\textwidth]{Figures/AllInsertions.pdf}
% q \caption{Our algorithm was evaluated with tight tolerance tasks, insertion toys, and objects from the NIST task board. }
%  \label{fig:allInsertions} 
% \end{figure}

\textbf{Rectangle Case Study: }With knowledge of the \textit{force plateaus}, we utilize this information to acquire and maintain contact formations. Leveraging our algorithm (Sec. \ref{sec_methods}), we set target modulating forces to 0.7N, 0.7N, and 1.5N along the $x-$, $y-$, and $z-axes$, respectively, which is $\sim$40-50\% of the maximum force before slippage occurs. 

In our first experiment, we attempt to insert the tight tolerance \textbf{(d)} rectangle with an initial 27.1\textdegree$ $ axial offset along the object's $z-axis$ (Fig. \ref{fig:forcePlot}(a)). Notably, the exact location of the hole is unknown, and the system is given a search space of 8cm$\times$8cm. The process begins by first finding a downward force, i.e., the first constraint, and choosing an exploration direction (in this case $+f_x$) within the search space until an additional constraint is detected (transition 3). The additional force spike on the $x-axis$ signals the hole's perimeter and a force transition begins. This process is evidenced in Fig. \ref{fig:forcePlot}(b), where forces are continually added and modulated around setpoints for states 1-7 (Sec. \ref{sec_methods}). Note that although compliant, the WAM robot has difficulty modulating velocities, and thus forces, due to controllability. Moreover, forces in Fig. \ref{fig:forcePlot}(b) are represented in the world frame to represent rotations during steps 4-5 and are smoothed for clarity, so force spikes are not represented.
% Ideally, $z-axis$ forces should be $\sim$1.5N between states 2-6. 

\textbf{All Objects: }Generalization is underscored by our ability to insert objects of varying convex, or near-convex, geometries -- specifically pegs with circle-, pear-, and triangle-shaped faces (Fig. \ref{fig:objs}). The result of our evaluations with tolerances, exploration times, insertion times, and offset degrees is presented in Table \ref{tab:main_insertion} \textbf{(a-d)}. During evaluation, we noted that the \textbf{(a)} circle was the easiest, as it did not require $z-axis$ offset control. The other objects \textbf{(b-d)}, were more difficult and posed various challenges. First, the initial offset was different for each, ranging from 27.1\textdegree $ $ to 41.1\textdegree. Objects (\textbf{c}) and (\textbf{d}) had sharp edges which encouraged jamming, whereas object (\textbf{b}) presented difficultly due to its non-convexity. Overcoming these challenges, we were able to complete insertions successfully with objects of $<$0.25mm tolerance and without prior knowledge of the hole pose (Fig. \ref{fig:forcePlot}(c)).

\subsection{Open World Insertion Tasks}
Beyond our tight tolerance evaluations, we were interested in applicability of our method to open world tasks. Our first experiment is with a commercially available children's toy consisting of different object geometries and with a hole tolerance of approximately 2.5mm. Similar to the previous experiments, the pose of the hole is unknown, and the search area is now confined to a 6.4cm$\times$6.4cm space (denoted in green in Fig. \ref{fig:objs}\textbf{(h)}). Here, we attempt to challenge the exploration component of our algorithm, ensuring the object started at the edge of the search space. All insertions were successfully completed, with the longest exploration phase of 63.2 seconds for the \textbf{(g)} clove. Interestingly, the non-convexity of the clove did not complicate the insertion process as much as originally believed, as the round edges of the object helped limit jamming from occurring. 

Our final experiment evaluated plug and gear insertion from the NIST Assembly Task Board (Fig. \ref{fig:forcePlot}(c)). The gear task is interesting in that the peg-in-hole paradigm is transformed instead into a hole-on-peg schema. This was not a problem for our system, as the search pattern was instead completed on the bottom of the hole, i.e., the gear, instead of by using a peg. Similarly, the plug insertion with a relaxed plunger spring was completed with ease, which started with an $z-axis$ offset of 36.2\textdegree$   $(Table \ref{tab:main_insertion}). These tests underscore the practicality and generalizability of our method, which is further showcased in the supplementary video.

\section{Discussions and Future Work}
This work presents a method that leverages \textit{compliance-enabled contact formations} as a step towards generalized, tight tolerance insertion for robots. Our algorithm is simple, yet mechanically grounded, and exploits the concept of reducing uncertainty through additive contact constraints, i.e., manipulation funnels. Notably, in this work we do not need to utilize costly learning frameworks or system-specific, idealized analytical models -- this method is effective yet did not require a single equation to describe in detail.  
% simple in nature. where we detect, acquire, and modulate constraints as desired to complete the insertion process.

The authors are excited about this preliminary exploration, as our experiments illustrate validity of our approach for future applications. As an attempt to not overclaim contributions, the authors want to be forthcoming on known limitations that will warrant future investigation:

\begin{enumerate}
  \item We cannot claim theoretical guarantees that an insertion will always be successful. In simulation, we verified that convex objects of non-negative tolerances should always succeed, but guarantees are not as clear for all non-convex circumstances.
  \item Hole geometry requires a low-frictional ``platform" for object exploration in order to find constraints. If this platform has variable friction, contact states may be transitioned prematurely and cause failure.
  \item Negative tolerance insertion would be unlikely due to the maximum forces the system can apply (Sec. \ref{sec:forcePlateaus}). A redesign of the end effector to apply a greater amount of force ($\sim$5-10N) would be beneficial for varied tasks. 
\end{enumerate}

Overall, the authors believe compliance will continue to prove invaluable for future advancements in robot manipulation. By investigating how to build more capable end effectors and by developing robust control strategies for non-convex object insertion, our method should prove to extend to a vast array of everyday insertion tasks for service robots of the future. Please find a complete overview of our motivation, approach, and experiments in the supplementary video.

% \section*{ACKNOWLEDGMENT}

\clearpage
\bibliographystyle{IEEEtran}
\bibliography{contact}
% \balance

\end{document}